\documentclass[10pt,twocolumn,letterpaper]{article}

\usepackage{cvpr}              

\usepackage{graphicx}
\usepackage{amsmath}
\usepackage{amssymb}
\usepackage{booktabs}
\usepackage[table]{xcolor}
\usepackage{float}

\newcommand{\our}{{CHASE}\xspace}
\newcommand{\myparagraph}[1]{\paragraph{#1}}

\definecolor{cvprblue}{rgb}{0.21,0.49,0.74}
\usepackage[pagebackref,breaklinks,colorlinks,allcolors=cvprblue]{hyperref}

\usepackage[capitalize]{cleveref}
\crefname{section}{Sec.}{Secs.}
\Crefname{section}{Section}{Sections}
\Crefname{table}{Table}{Tables}
\crefname{table}{Tab.}{Tabs.}

\def\cvprPaperID{7806} 
\def\confName{CVPR}
\def\confYear{2025}

\title{CHASE: 3D-Consistent Human Avatars with Sparse Inputs via Gaussian Splatting and Contrastive Learning}

\author {
    Haoyu Zhao\textsuperscript{* \rm 1,\rm 2},
    Hao Wang\textsuperscript{* \rm 3},
    Chen Yang\textsuperscript{* \rm 1},
    Wei Shen\textsuperscript{\textdagger \rm 1} \\
    \textsuperscript{1}MoE Key Lab of Artificial Intelligence, AI Institute, Shanghai Jiao Tong University \quad \\
    \textsuperscript{2}School of Computer Science, Wuhan University \quad \\
    \textsuperscript{3}Wuhan National Laboratory for Optoelectronics, Huazhong University of Science and Technology \quad
}

\begin{document}

\twocolumn[{%
    \renewcommand\twocolumn[1][]{#1}%
    \setlength{\tabcolsep}{0.0mm} 
    \newcommand{\sz}{0.125}  
    \maketitle
    \begin{center}
        \newcommand{\teaserwidth}{\textwidth}
        \includegraphics[width=\linewidth]{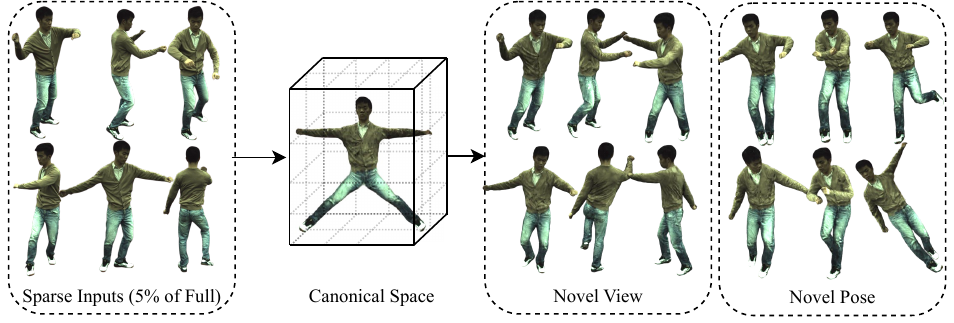}
    \captionof{figure}{\textbf{\our.} We propose an efficient method for creating 3D-consistent animatable avatars from just videos. Our method achieve better quality to the most recent SOTA methods~\cite{wen2024gomavatar,hu2024gauhuman,qian20243dgs} in both full and sparse inputs.} 
    \label{fig:teaser}
    \end{center}%
}]

\maketitle
{
\renewcommand{\thefootnote}{\fnsymbol{footnote}}
\footnotetext{* Equal contributions.}
\footnotetext{\textdagger Corresponding Author.}
\footnotetext{Haoyu Zhao completed this work during an internship at Shanghai Jiao Tong University.}
}
\maketitle

\begin{abstract}
Existing approaches for human avatar generation--both NeRF-based and 3D Gaussian Splatting (3DGS) based--struggle with maintaining 3D consistency and exhibit degraded detail reconstruction, particularly when training with sparse inputs.
To address this challenge, we propose \textbf{\our}, a novel framework that achieves dense-input-level performance using only sparse inputs through two key innovations: cross-pose intrinsic 3D consistency supervision and 3D geometry contrastive learning. Building upon prior skeleton-driven approaches that combine rigid deformation with non-rigid cloth dynamics, we first establish baseline avatars with fundamental 3D consistency.
To enhance 3D consistency under sparse inputs, we introduce a Dynamic Avatar Adjustment (DAA) module, which refines deformed Gaussians by leveraging similar poses from the training set. By minimizing the rendering discrepancy between adjusted Gaussians and reference poses, DAA provides additional supervision for avatar reconstruction. We further maintain global 3D consistency through a novel geometry-aware contrastive learning strategy. While designed for sparse inputs, \our surpasses state-of-the-art methods across both full and sparse settings on ZJU-MoCap and H36M datasets, demonstrating that our enhanced 3D consistency leads to superior rendering quality. Project page: 
\href{https://chaseprojectpage.github.io/}{https://chaseprojectpage.github.io/}.
\end{abstract}

\section{Introduction}
\label{sec:intro}
Photo-realistic rendering and animation of human bodies is a critical research area with wide-ranging applications in AR/VR, visual effects, virtual try-on, and film production~\cite{healey2021mixed}. Early approaches~\cite{niemeyer2020differentiable} relied on multi-camera setups to capture high-quality data, requiring extensive computational resources and manual effort. While these methods perform well for reconstructing a single scene or object with sufficient input views, they struggle to generalize to new scenes or objects from limited samples~\cite{kwon2024generalizable}.



Recent advancements have explored using neural radiance fields (NeRF) for modeling 3D human avatars~\cite{mildenhall2021nerf}, typically employing parametric body models to model deformations. Some methods~\cite{chen2022geometry,zhao2022humannerf,jiang2022instantavatar} use human template models to facilitate generalizable and robust synthesis. However, NeRF-based methods are less efficient to train and render due to their computationally intensive per-pixel volume rendering process. 

Point-based rendering~\cite{zheng2023pointavatar} has emerged as an efficient alternative to NeRFs, offering significantly faster rendering. 
The recently proposed 3D Gaussian Splatting (3DGS)~\cite{kerbl20233d} gains popularity for its fast rendering speed. Numerous works have further explored the 3D Gaussian representation for dynamic 3D human avatars~\cite{lei2024gart,moreau2024human,shao2024splattingavatar,hu2024gauhuman,kocabas2024hugs,wang2024gaussian,qian20243dgs,hu2024gaussianavatar}. However, these methods often face challenges in maintaining 3D consistency and producing high-quality reconstructions, particularly with sparse inputs.

To address the aforementioned issues, we propose \textbf{\our}, which is capable of reconstructing 3D \textbf{C}onsistent \textbf{H}uman \textbf{A}vatars with \textbf{S}parse inputs via Gaussian Splatting and contrastiv\textbf{E} learning, as shown in Fig.~\ref{fig:teaser}.
We first integrate a skeleton-driven rigid deformation and a non-rigid cloth dynamics deformation to create a human avatar.
To enhance 3D consistency under sparse inputs, we utilize the intrinsic 3D consistency of images across different poses within the same person. Specifically, for each training pose/image, we select a similar pose/image from the dataset and then adjust the deformed Gaussians using the proposed Dynamic Avatar Adjustment (DAA), an explicit point-based control graph adjustment strategy, to the selected similar pose. 
Then, we minimize differences between the rendered image of the adjusted Gaussians and the image corresponding to the selected similar pose, which serves as an additional form of supervision for human avatars.
Additionally, we employ 3D geometry contrastive learning, utilizing features from a 3D feature extractor, to further enhance the global 3D consistency of generated human avatars.
We conduct extensive experiments on the  ZJU-MoCap data~\cite{peng2020neural}, H36M~\cite{ionescu2013human3} and surprisingly find that \textit{\our outperforms other SOTAs in both full and sparse inputs setting.} 
Our work makes the following contributions:
\begin{itemize}
    \item We propose an explicit point-based control graph adjustment strategy, which introduces a novel 2D image supervision to 3D human body modeling, enhancing the 3D consistency of human avatars.
    \item We propose a 3D geometry contrastive learning to enforce consistency across different representations of the same pose and enhance global 3D understanding.
    \item Extensive experiments show that our \our achieves SOTA performance quantitatively and qualitatively under full and sparse settings.
\end{itemize}


\section{Related Work}
\subsection{Contrastive Representation Learning}


Contrastive Representation Learning is one of the mainstream self-supervised learning paradigms, which learns potential semantics from constructed invariance or equivariance. In 3D, PointContrast~\cite{xie2020pointcontrast} proposes geometric augmentation to generate positive and negative pairs. CrossPoint~\cite{afham2022crosspoint} uses both inter- and intra-modal contrastive learning. PointCLIP~\cite{zhang2022pointclip} achieves image-point alignment by projecting point clouds onto 2D depth images. RECON~\cite{qi2023contrast} focuses on single- and cross-modal contrastive learning through discriminative contrast~\cite{khosla2020supervised} or global feature alignment~\cite{radford2021learning}. Our \our introduces a 3D geometry contrastive learning method to enforce consistency across different representations of the same pose.

\subsection{3D Editing and Deformation}
Traditional deformation methods in computer graphics are typically based on Laplacian coordinates~\cite{gao2019sparse}, Poisson equations~\cite{yu2004mesh}, and cage-based methods~\cite{yifan2020neural}. However, these methods often rely on implicit and computationally expensive NeRF-based approaches.

Numerous works~\cite{chen2024gaussianeditor,zhao2024hfgs} have proposed techniques for editing 3D Gaussian Splatting (3DGS)\cite{kerbl20233d}. SuGaR\cite{guedon2024sugar} introduces a mesh extraction method that produces meshes from 3DGS, which can then be edited. SC-GS~\cite{huang2024sc} proposes deforming Gaussians by transferring the movement of control points. Our \our employs a novel explicit point-based control graph deformation strategy, which is more efficient than previous methods.

\subsection{3D Human Modeling}
Since the high-quality rendering achieved by the seminal work Neural Radiance Fields (NeRF)\cite{mildenhall2021nerf}, there has been a surge of research on neural rendering for human avatars\cite{liu2021neural, li2023posevocab, peng2020neural,remelli2022drivable}. Although NeRF is designed for static objects, HumanNeRF~\cite{weng2022humannerf} extends NeRF to enable capturing dynamic human motion using just a single monocular video. Neural Body~\cite{peng2020neural} associates a latent code with each SMPL~\cite{loper2015smpl} vertex to encode appearance, which is then transformed into observation space based on the human pose. Furthermore, Neural Actor~\cite{liu2021neural} learns a deformable radiance field with SMPL~\cite{loper2015smpl} as guidance and utilizes a texture map to improve the final rendering quality. Posevocab~\cite{li2023posevocab} designs joint-structured pose embeddings to encode dynamic appearances under different key poses, allowing for more effective learning of joint-related appearances. However, a major limitation of NeRF-based methods is that NeRFs are slow to train and render.

Point-based rendering~\cite{zheng2023pointavatar} has proven to be an efficient alternative to NeRFs for fast inference and training. Extending point clouds to 3D Gaussians, 3D Gaussian Splatting (3DGS)\cite{kerbl20233d} models the rendering process by splatting a set of 3D Gaussians onto the image plane via alpha blending. Given the impressive performance of 3DGS in both quality and speed, numerous works have further explored the 3D Gaussian representation for dynamic 3D human avatar reconstruction\cite{lei2024gart, qian20243dgs, hu2024gauhuman, kocabas2024hugs}. Human Gaussian Splatting~\cite{moreau2024human} showcases 3DGS as an efficient alternative to NeRF. SplattingAvatar~\cite{shao2024splattingavatar} and GomAvatar~\cite{wen2024gomavatar} extend lifted optimization to simultaneously optimize the parameters of the Gaussians while walking on the triangle mesh. However, these methods struggle to maintain 3D consistency and produce low-quality reconstructions when applied to human avatar creation with only sparse inputs. Our \our introduces a novel 2D image supervision to 3D human body modeling and 3D geometry contrastive learning, enhancing the 3D consistency of human avatars.

\begin{figure*}[!htb]
  \centering
    \includegraphics[width=\linewidth]{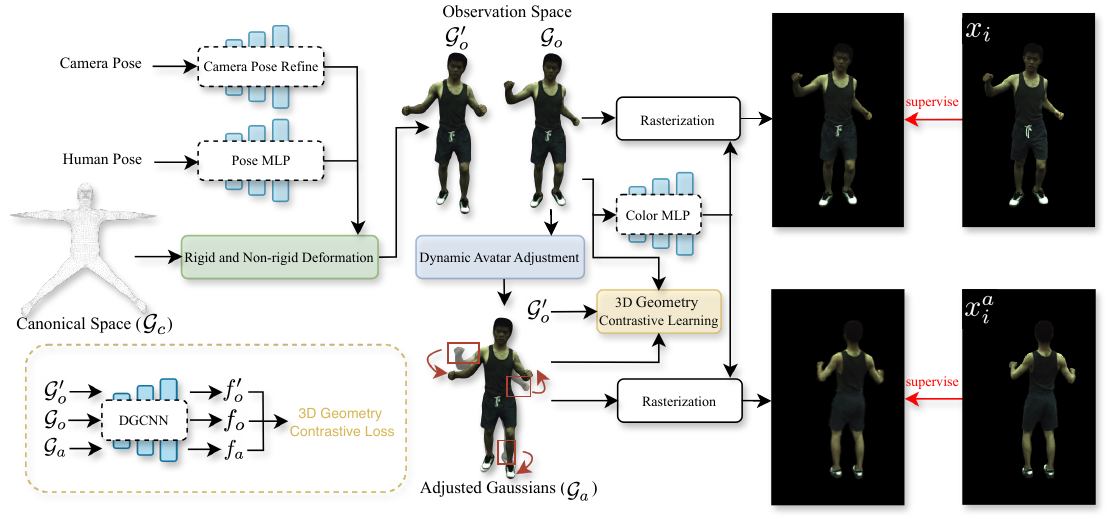}\\
  \caption{\textbf{\our Framework.} We first initialize 3D Gaussians in canonical space by randomly sampling 50k points on the SMPL mesh surface. Then, we integrate a rigid human articulation and a non-rigid deformation neural field to deform the 3D Gaussians in canonical space ${\mathcal{G}_c}$ to the observation space ${\mathcal{G}_o}$. Next, we select similar poses/images from the dataset for each training pose/image and then adjust the deformed Gaussians ${\mathcal{G}_o}$ to the similar pose ${\mathcal{G}_a}$ using Dynamic Avatar Adjustment (DAA). Minimizing the differences between the rendered adjusted Gaussians ${\mathcal{G}_a}$ and the selected similar images $x_i^a$ serves as an additional supervision. Furthermore, we propose a 3D geometry contrastive learning, which involves comparing features from a 3D feature extractor to improve the avatar’s global 3D consistency. Negative pairs consist of the features of the deformed Gaussians ${\mathcal{G}_o}$ and the adjusted Gaussians ${\mathcal{G}_a}$. In contrast, positive pairs include the features of ${\mathcal{G}_o'}$, which is deformed from the canonical space to match the pose adjustments seen in ${\mathcal{G}_a}$, and ${\mathcal{G}_a}$.}
  \label{fig:pipeline}
\end{figure*}


\section{Preliminaries}

\medskip
\noindent
\textbf{SMPL~\cite{loper2015smpl}.} The SMPL model is a pre-trained parametric human model representing body shape and pose. In SMPL, body shape and pose are controlled by pose and shape. In this work, we apply the Linear Blend Skinning (LBS) algorithm used in SMPL to transform points from a canonical space to a posed space.

\medskip
\noindent
\textbf{LBS~\cite{sumner2007embedded}.} Linear Blend Skinning (LBS) is a weight-based technique that associates each vertex with one or more joints and uses weight values to describe the influence of each joint on the vertex. Vertex deformation is calculated by linearly interpolating transformations on the associated joints: $\mathcal{X}_{v}^{'} = \sum_{j=1}^{J}w_j(\mathcal{X}_{v})B_j\mathcal{X}_{v}$, where $J$ represents the number of joints, $N$ represents the number of vertices,
$\mathcal{X}_{v}^{'} \in \mathbb{R}^{N\times3}$ is the new position of the skinned vertex, $w \in \mathbb{R}^{N \times J}$ is the skinning weight matrix, $ B\in \mathbb{R}^{J\times 4\times 4}$ is the affine transformation matrix of each joint representing rotation and translation, (i.e. bone transforms) and $\mathcal{X}_{v}\in \mathbb{R}^{N\times3}$ is the original mesh vertex position.

\medskip
\noindent
\textbf{3D Gaussian Splatting (3DGS)~\cite{kerbl20233d}.} 3DGS explicitly represents scenes using point clouds, where each point is modeled as a 3D Gaussian defined by a covariance matrix $\Sigma$ and a center point $\mathcal{X}$, the latter referred to as the mean. The value at point $\mathcal{X}$ is: 

\begin{equation}
G(\mathcal{X})=e^{-\frac{1}{2}\mathcal{X}^T\Sigma^{-1}\mathcal{X}}.
\end{equation}
For differentiable optimization, the covariance matrix $\Sigma$ is decomposed into a scaling matrix $\mathcal{S}$ and a rotation matrix $\mathcal{R}$, such that $\Sigma = \mathcal{R}\mathcal{S}\mathcal{S}^T\mathcal{R}^T$. $\mathcal{S}$ and $\mathcal{R}$ are stored as the diagonal vector $s\in \mathbb{R}^{N\times3}$ and a quaternion vector $r\in \mathbb{R}^{N\times4}$, respectively.

In rendering novel views, differential splatting as introduced by~\cite{yifan2019differentiable}, involves using a viewing transform $W$ and the Jacobian matrix $J$ of the affine approximation of the projective transformation to compute the transformed covariance matrix: $\Sigma^{\prime} = JW\Sigma W^TJ^T$. 
The color and opacity at each pixel are computed from the Gaussian's representations: $G(\mathcal{X})=e^{-\frac{1}{2}\mathcal{X}^T\Sigma^{-1}\mathcal{X}}.$ The blending of $N$ ordered points overlapping a pixel is given by the formula: $\mathcal{C} = \sum_{i\in N}c_i \alpha_i \prod_{j=1}^{i-1} (1-\alpha_i),$ where $c_i$, $\alpha_i$ represent the density and color of this point computed by a 3D Gaussian $G$ with covariance $\Sigma$ multiplied by an optimizable per-point opacity and SH color coefficients. 


\section{Method}
We illustrate the pipeline of our \our in Fig.~\ref{fig:pipeline}. The inputs include images $X = \{x_i\}^N_{i=1}$ obtained from monocular videos, fitted SMPL parameters $P = \{p_i\}^N_{i=1}$, and foreground masks $M = \{m_i\}^N_{i=1}$ of images. \our optimizes 3D Gaussians in canonical space, which are then be deformed to match the observation space and be rendered with a given camera view. 


\subsection{Non-rigid and Rigid Deformation}
\label{sec: deformation}
Inspired by~\cite{weng2022humannerf,qian20243dgs}, we deform 3D Gaussians from canonical space ${\mathcal{G}_c}$ to observation space ${\mathcal{G}_o}$ by integrating a rigid articulation with a non-rigid transformation. We employ a non-rigid deformation network that takes the canonical positions $\mathcal{X}_c$ of the 3D Gaussians ${\mathcal{G}c}$ and a pose latent code which encodes SMPL pose $p_i$ using a lightweight hierarchical pose encoder~\cite{mihajlovic2021leap}. The network then outputs the offsets for various parameters of the 3D Gaussians $\mathcal{G}_c$: $\Delta {(\mathcal{X}, \mathcal{C}, \alpha, s, r)}$. The canonical Gaussians are deformed by:
\begin{align}
\mathcal{X}_d &= \mathcal{X}_c + \Delta \mathcal{X}, \mathcal{C}_d = \mathcal{C}_c + \Delta \mathcal{C} \label{eq:deform1}, \\
\alpha_d &= \alpha_c + \Delta \alpha, s_d = s_c \cdot\exp (\Delta s), \label{eq:deform2} \\
r_d &= r_c \cdot [1,\Delta r_1,\Delta r_2, \Delta r_3], \label{eq:deform3}
\end{align}
where quaternion multiplication $\cdot$ corresponds to multiplying the rotation matrices. With $[1,0,0,0]$ as the identity rotation, $r_d = r_c$ when $\delta {r} = \mathbf{0}$, thus keeping the original orientation.

We further apply a LBS-based rigid transformation to map the non-rigidly deformed 3D Gaussians ${\mathcal{G}_d}$ to the observation space ${\mathcal{G}_o}$. This transformation utilizes LBS weights predicted by a Skinning MLP $f_{\theta_r}$. This process aligns the Gaussians with the target pose in ${\mathcal{G}_o}$:
\begin{align}
    &T = \sum_{j=1}^J f_{\theta_r}(\mathcal{X}_d)_j B_j,  
    \mathcal{X}_o = T\mathcal{X}_d, \\
    &\mathcal{R}_o = T_{1:3,1:3}\mathcal{R}_d,
\end{align}
where $\mathcal{R}$ is the matrix representations of rotation. 


\begin{figure}[t]
  \centering
    \includegraphics[width=\linewidth]{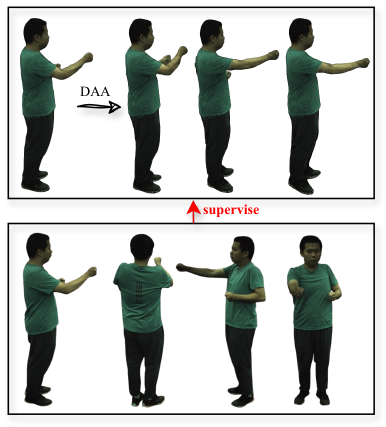}\\

  \caption{For each training pose/image, we select similar poses/images from the dataset and then adjust the deformed Gaussians using DAA. By minimizing the difference between the rendered image of the adjusted avatar and the selected similar pose image, we introduce additional supervision, thereby refining the creation of photo-realistic and animatable avatars.}
  \label{fig:daa}
\end{figure}

\subsection{Dynamic Avatar Adjustment}
\label{sec:daa}
To address extremely sparse inputs, we leverage the intrinsic 3D consistency of human avatars across different poses/images, as shown in Fig.~\ref{fig:daa}. Though the same pose may exhibit slight variations in non-rigid deformation, these differences occupy only a small number of pixels from a whole-body perspective and thus have minimal impact on the overall body reconstruction, which we further demonstrate in the experimental Section.~\ref{sec:ablation}. 

Specifically, for each training pose/image, we select a similar pose $p^a_i$ with its paired image $x_i^a$ by computing the orientation and limb angle difference provided by SMPL~\cite{loper2015smpl} model from the dataset. Then we use a dense motion field $F_{adj}$ as an additional adjustment to transform deformed Gaussians ${\mathcal{G}_o}$ into adjusted Gaussians ${\mathcal{G}_a}$, aligning them with the selected pose/image ($p_i^a$/$x_i^a$). In this way, we successfully introduce an additional 2D image supervision, improving the 3D consistency of human avatars.

To achieve precise control of the 3D Gaussians, we sample 6,890 points from the SMPL model~\cite{loper2015smpl} as our sparse control points in canonical space. Then, we obtain the dense motion field using LBS by locally inheriting the LBS weights from neighboring control points. Specifically, for each 3D Gaussian, we use the k-nearest neighbor (KNN) search to find its nearest neighboring control points in canonical space. The entire adjustment process is as:
\begin{align}
w=w_{smpl}[\text{KNN}(xyz_{cano},xyz_{smpl})],\\
T_{o}=\sum_{j=1}^{J}w_jB_{oj}, T_{o}^{'}=\sum_{j=1}^{J}w_jB_{oj}^{'}.
\end{align}
Here, $w_{smpl}$ denotes the LBS weights of the sparse control points, and $T_{o}(B_{o})$ and $T_{o}'(B_{o}')$ represent the rigid transformations (bone transformations) from canonical space $\mathcal{G}_c$ to deformed Gaussians $\mathcal{G}_o$, and to the Gaussians with the selected similar pose $\mathcal{G}_o'$, respectively. We then obtain $F_{adj}$, which transforms the deformed Gaussians $\mathcal{G}_o$ into adjusted Gaussians $\mathcal{G}_a$, aligning them with the selected pose $p_i^a$ as:
\begin{align}
F_{adj}&=F_{o'}F_{o}^{-1}.
\end{align}
We adjust the deformed Gaussians ${\mathcal{G}_o}$ to adjusted Gaussians ${\mathcal{G}_a}$ by by adjusting its position and rotation as follow:
\begin{align}
\mathcal{X}_{a}&=F_{adj}\mathcal{X}_{o},\\
\mathcal{R}_{a}&=F_{adj 1:3,1:3}\mathcal{R}_{o}.
\end{align}





\subsection{3D Geometry Contrastive learning}
Inspired by the success of contrastive learning in 2D image processing and static point cloud analysis, we advocate for adopting 3D geometry contrastive learning to ensure 3D consistency of avatars. We treat the 3D Gaussians as a 3D point cloud and use DGCNN~\cite{wang2019dynamic} as the feature extractor. DGCNN is typically trained on general point cloud datasets, which helps it learn geometric structures. The point cloud feature extractor processes the positions of the 3D Gaussians in the observation space ${\mathcal{G}_o}$, the adjusted Gaussians ${\mathcal{G}_a}$, and ${\mathcal{G}_o'}$, which is deformed from the canonical space to match the selected pose $p_i^a$, and outputs their features, creating intermediate graph features to capture global geometric information better. The feature vectors are projected into an invariant space. We denote the projected features of ${\mathcal{G}_o}$, ${\mathcal{G}_a}$, and ${\mathcal{G}_o'}$ as $f_o$, $f_a$, and $f_o'$, respectively.

In the invariant space, we aim to maximize the similarity between $f_a$ and $f_o'$, denoted as $\text{D}_{positive}$, and minimize the similarity between $f_a$ and $f_o$, denoted as $\text{D}_{negative}$. This feature-level contrastive learning encourages the model to capture pose-aware knowledge, enabling it to differentiate between subtle pose variations while maintaining geometric consistency. Therefore, we compute the 3D geometry contrastive loss $\mathcal{L}_{contrastive}$ as:
\begin{align}
    \text{D}_{positive} &= \|f_a - f_o'\|_2, \\
    \text{D}_{negative} &= \|f_a - f_o\|_2, \\
    \mathcal{L}_{contrastive} = \max(0  &,  \text{D}_{positive} - \text{D}_{negative}).
\end{align}


\subsection{Optimization}
We begin by randomly sampling 50k points from the surface of the SMPL mesh to initialize 3D Gaussians in the canonical space.
\myparagraph{Color MLP.}
Following~\cite{qian20243dgs}, we use the inverse rigid transformation to canonicalize the viewing direction: $\hat{d}=T_{1:3,1:3}^{-1}d$, where $T$ and $d$ is the forward transformation matrix defined in LBS and viewing direction, respectively. Theoretically, canonicalizing viewing direction also promotes consistency of the specular component of canonical 3D Gaussians under rigid transformations.

\myparagraph{Pose correction.} 
Following~\cite{qian20243dgs}, SMPL~\cite{loper2015smpl} parameter fittings from images can be inaccurate. We additionally optimize the per-sequence shape parameter and per-frame translation, global rotation, and local joint rotations.

\myparagraph{Loss function.}
Our full loss function consists of several components: an RGB loss $\mathcal{L}_{rgb}$, a mask loss $\mathcal{L}_{mask}$, and a perceptual similarity (LPIPS) loss $\mathcal{L}_{LPIPS}$. We compute these losses on both images rendered from the deformed Gaussians $\mathcal{G}_o$ and the adjusted Gaussians $\mathcal{G}_a$ with their corresponding ground truth images.
Additionally, we include a skinning weight regularization loss $\mathcal{L}_{skin}$, as well as isometric regularization losses for both position and covariance, $\mathcal{L}_{isopos}$ and $\mathcal{L}_{isocov}$, following~\cite{qian20243dgs}.
We also incorporate a 3D geometry contrastive loss $\mathcal{L}_{contrastive}$:
\begin{align}
\mathcal{L}=  &\mathcal{L}_{rgb} + \lambda_1 \mathcal{L}_{mask} + \lambda_2 \mathcal{L}_{LPIPS} + \lambda_3 \mathcal{L}_{skin} + \nonumber\\
 &  \lambda_4 \mathcal{L}_{isopos} + \lambda_5 \mathcal{L}_{isocov} + \lambda_6 \mathcal{L}_{contrastive},
\label{eq:loss}
\end{align}
where $\lambda$'s are loss weights. For further details of the loss definition and respective weights, please refer to the Supp.Mat.


\begin{figure*}[!htb]
  \centering
    \includegraphics[width=\linewidth]{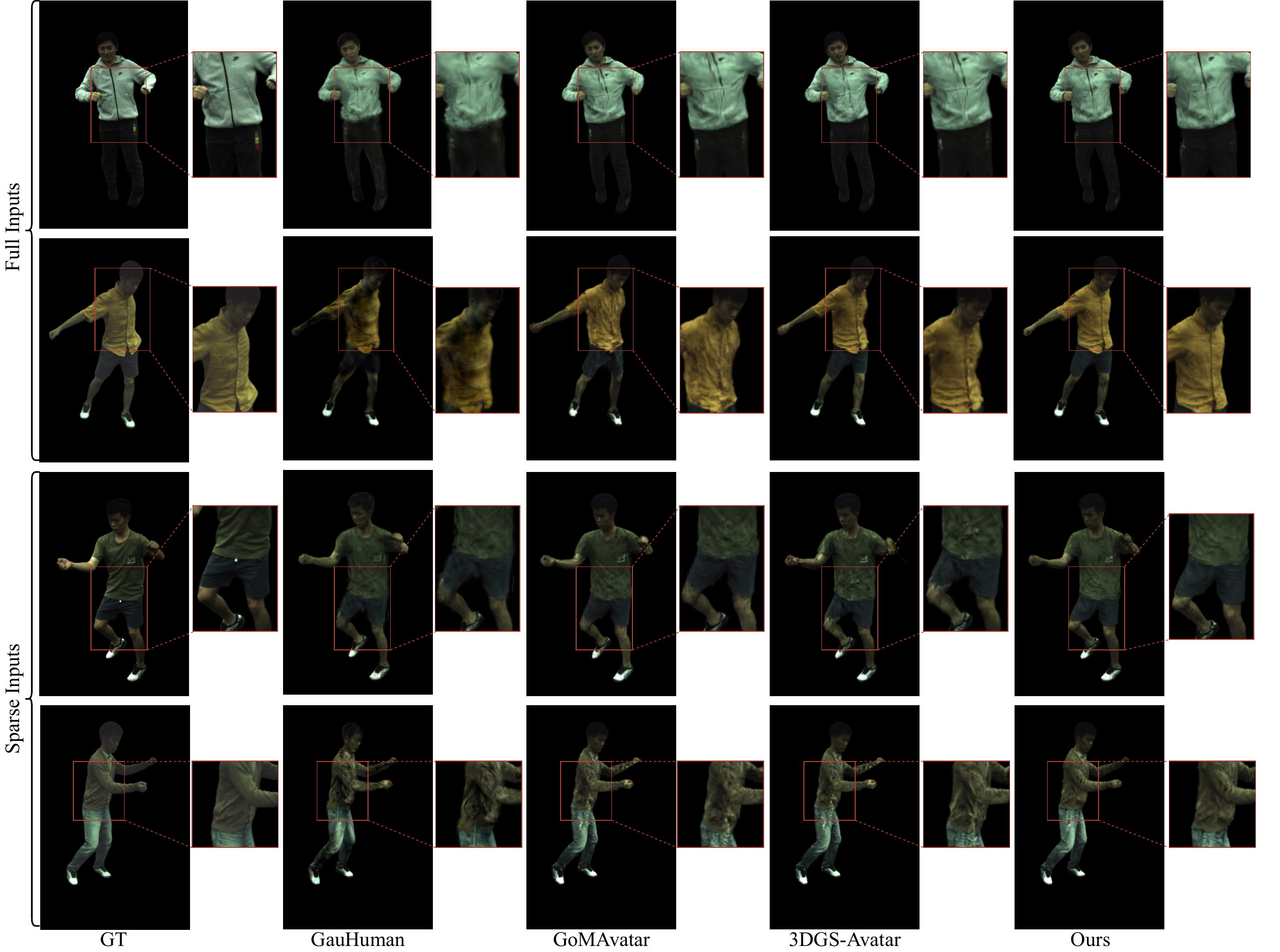}\\

  \caption{\textbf{Qualitative Comparison on ZJU-MoCap~\cite{peng2020neural}.} We present results for full and sparse inputs (5\% of the full inputs) on the ZJU-MoCap dataset. Results show that our \our can produce realistic details with both full and sparse inputs, while other approaches struggle to generate smooth details.}
  \label{fig:result}
\end{figure*}


\begin{figure}[!htb]
  \centering
    \includegraphics[width=\linewidth]{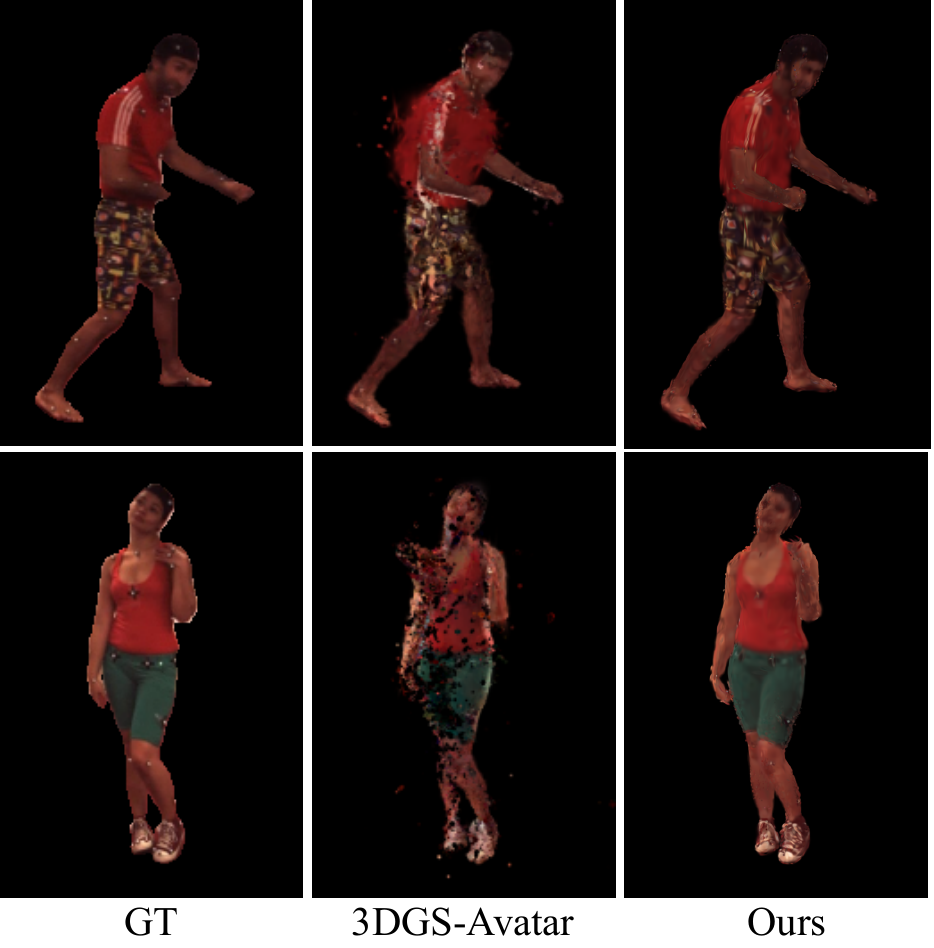}\\
  \caption{\textbf{Qualitative Comparison on H36M~\cite{ionescu2013human3} with sparse inputs.} We demonstrate that our method effectively produces realistic details for novel pose in both rendered images and geometry, whereas other approach struggles to achieve smooth details.}
  \label{fig:result_h36m}
\end{figure}

\begin{table}[t]
\setlength{\fboxsep}{2pt}
\fontsize{9}{10}\selectfont
 \caption{\textbf{Quantitative Results on ZJU-MoCap~\cite{peng2020neural}.} \our achieves state-of-the-art performance across every method. The \colorbox{pink}{best} and the \colorbox{yellow}{second best} results are denoted by pink and yellow. Frames per second (FPS) is measured on an RTX 3090. We train our model on the dataset that includes only 5\% of the origin data for fair quantitative comparison. The metrics are reported in the last four rows of the table. LPIPS$^{\dagger}$ = LPIPS $\times$ 1000.}
 \label{tab:compare_zjumocap}
 \centering
 \setlength{\tabcolsep}{2pt}
 \renewcommand{\arraystretch}{1.1}
 \begin{tabular}{ l|cccc}
 \toprule
Method:     
& LPIPS$^{\dagger}$$\downarrow$ 
& PSNR$\uparrow$
& SSIM$\uparrow$   
& FPS 
 \\\hline
 NeuralBody~\cite{peng2020neural}  & 52.29 & 29.07 & 0.962  & 1.5\\ 
 Ani-NeRF~\cite{peng2021animatable} & 51.98 & 29.17 & 0.961 & 1.1\\
 HumanNeRF~\cite{weng2022humannerf} & 31.73 & 30.24 & 0.968 & 0.3\\
 MonoHuman~\cite{yu2023monohuman} & 37.51 & 29.38 & 0.964 & 0.1 \\
 DVA~\cite{remelli2022drivable} & 37.74 & 29.45 & 0.956 & 17 \\
 InstantAvatar~\cite{jiang2022instantavatar} & 64.41 & 29.73 & 0.938 & 4.2\\
 3DGS-Avatar~\cite{qian20243dgs}   & \cellcolor{yellow}30.28 & 30.62 & 0.965 & \cellcolor{yellow}50\\
 GauHuman~\cite{hu2024gauhuman} & 32.73 & \cellcolor{yellow}30.79 & 0.960 & \cellcolor{pink}180\\ 
 GoMAvatar~\cite{wen2024gomavatar} & 32.53 & 30.37 & \cellcolor{yellow}0.969 & 43 \\
 Ours & \cellcolor{pink}27.48 & \cellcolor{pink}30.81 & \cellcolor{pink}0.970 & \cellcolor{yellow}50\\
\hline
 3DGS-Avatar*~\cite{qian20243dgs}  & 40.01 & 29.98 & 0.957 & \cellcolor{yellow}50\\ 
 GauHuman*~\cite{hu2024gauhuman} & \cellcolor{yellow}35.68 & \cellcolor{yellow}30.35 & 0.957  & \cellcolor{pink}180\\ 
 GoMAvatar*~\cite{wen2024gomavatar} & 42.88 & 30.01 & \cellcolor{yellow}0.958 & 43\\ 
 Ours* & \cellcolor{pink}29.94 & \cellcolor{pink}30.48 & \cellcolor{pink}0.969 & \cellcolor{yellow}50 \\ 
 
 \bottomrule
 \end{tabular}
\end{table}


\section{Experiment}

\subsection{Dataset}

\medskip
\noindent
\textbf{ZJU-MoCap~\cite{peng2020neural}.}
This dataset features multi-view videos captured by 21 cameras, with human poses recorded using a marker-less motion capture system. For our experiments, we selected six sequences (377, 386, 387, 392, 393, 394). Following the protocol established by HumanNeRF~\cite{weng2022humannerf} and 3DGS-Avatar~\cite{qian20243dgs}, we use a single camera for training and the remaining cameras for evaluation. The foreground masks, camera, and SMPL parameters provided by the data set are used for evaluation purposes. \textit{We simulate sparse inputs by sampling every 20th frame from the video, between start\_frame and end\_frame, which corresponds to using 5\% of the total frames}. We also specify in the Supp.Mat which images are selected for training.

\medskip
\noindent
\textbf{H36M~\cite{ionescu2013human3}.}
H36M is another widely used dataset for human avatar research, comprising multi-view videos from four cameras and human poses captured via a marker-based motion capture system. We conducted experiments on sequences from subjects S1, S5, S6, S7, S8, S9, and S11, selecting representative actions and dividing the videos into training and test frames. Adhering to the protocol set by ARAH~\cite{ARAH:ECCV:2022}, we use three cameras, [54138969, 55011271, 58860488], for training and the remaining camera, [60457274], for testing, and follow their preprocessing steps. We use the SMPL parameters and foreground humans following~\cite{peng2021animatable}. We apply the same approach as mentioned above to simulate sparse inputs in this dataset.


\begin{table}
\setlength{\fboxsep}{2pt}
\fontsize{9}{10}\selectfont
 \caption{\textbf{Quantitative Results on H36M~\cite{ionescu2013human3}}. Our \our outperforms current SOTA methods in both full and sparse settings}
 \label{tab:compare_h36m}
 \centering
 \setlength{\tabcolsep}{1pt}
 \renewcommand{\arraystretch}{1.1}
 \begin{tabular}{ l|cc|cc}
 \toprule                  
 & \multicolumn{2}{c}{Training Poses}                                 
 & \multicolumn{2}{c}{Novel Poses}                                 
 \\ 
Method:             
& PSNR$\uparrow$
& SSIM$\uparrow$   
& PSNR$\uparrow$
& SSIM$\uparrow$  
 \\\hline
NARF~\cite{noguchi2021neural} & 23.00 & 0.898 & 22.27 & 0.881 \\
NeuralBody~\cite{peng2020neural} & 22.89 & 0.896 & 23.09 & 0.891 \\
Ani-NeRF~\cite{peng2021animatable} & 23.00 & 0.890 & 22.55 & 0.880 \\
ARAH~\cite{ARAH:ECCV:2022} & 24.79 & 0.918 & 23.42 & 0.896  \\
3DGS-Avatar~\cite{qian20243dgs} & \cellcolor{yellow}32.89 & \cellcolor{yellow}0.982 & \cellcolor{yellow}32.50 & \cellcolor{pink}0.983  \\
 Ours & \cellcolor{pink}33.29 & \cellcolor{pink}0.984 & \cellcolor{pink}32.93 & \cellcolor{yellow}0.982     \\
 \hline
3DGS-Avatar*~\cite{qian20243dgs} & \cellcolor{yellow}32.48 & \cellcolor{yellow}0.976 & \cellcolor{yellow}32.17 & \cellcolor{yellow}0.981  \\
 Ours* & \cellcolor{pink}32.91 & \cellcolor{pink}0.983 & \cellcolor{pink}32.64 &  \cellcolor{pink}0.982    \\
 \bottomrule
  \end{tabular}
\end{table}

\subsection{Comparison with State-of-the-art Methods}

We compare our \our with various SOTA methods for human avatars, including NerF-based methods such as NeuralBody~\cite{peng2020neural}, Ani-NeRF~\cite{peng2021animatable}, HumanNeRF~\cite{weng2022humannerf}, and MonoHuman~\cite{yu2023monohuman}, and 3DGS-based methods such as 3DGS-Avatar~\cite{qian20243dgs}, GauHuman~\cite{hu2024gauhuman}, and GoMAvatar~\cite{wen2024gomavatar} under monocular setup on ZJU-MoCap~\cite{peng2020neural}. The quantitative results are shown in Tab.~\ref{tab:compare_zjumocap}. Overall, our proposed \our achieves the best performance in terms of PSNR, SSIM, and LPIPS \textit{with both full and sparse inputs}. Notably, \textit{our \our shows only a small performance drop when using only 5\% of the data}, especially in LPIPS which is more informative than the other two metrics in our setting~\cite{qian20243dgs}. In fact, it declines less compared to other methods and even surpasses their performance with 100\% of the data. It is evidence that our method successfully maintains 3D consistency even with sparse inputs. 

Qualitative comparisons on novel view synthesis are shown in Fig.~\ref{fig:result}. 
We observe that our method preserves more details compared to other SOTA methods. They often struggle to maintain 3D consistency and deliver suboptimal detail reconstruction in human avatar modeling, particularly when only sparse inputs are available. Please see our project website and supplementary material for more visualization.

For H36M~\cite{ionescu2013human3}, we report the quantitative results against NeRF-based methods such as NARF~\cite{noguchi2021neural}, NeuralBody~\cite{peng2020neural}, Ani-NeRF~\cite{peng2021animatable} and ARAH~\cite{ARAH:ECCV:2022}, and 3DGS-based methods such as 3DGS-Avatar~\cite{qian20243dgs} in Tab.~\ref{tab:compare_h36m}. Our \our significantly outperforms these methods with sparse inputs, showing that our \our generalizes well to novel poses with sparse inputs and reconstructs human avatars with better appearance and geometry detail. For qualitative comparisons on novel pose synthesis, as shown in Fig.~\ref{fig:result_h36m}, our method generalizes well to novel pose with just sparse inputs (only 5\% of origin data) and reconstruct human avatars with better appearance and geometry
detail.


\begin{table}
 \caption{\textbf{Ablation Study on ZJU-MoCap~\cite{peng2020neural}.}
  We both show the result from full input (top group) and sparse input (bottom group).}
 \label{tab:ablation}
 \centering
 \begin{tabular}{@{}lcccc}
 \toprule
 Method:          
 & PSNR$\uparrow$
 & SSIM$\uparrow$
 & LPIPS$^{\dagger}$$\downarrow$
 & FPS\\ 
 \hline
 w/o non-rigid & 30.32 & \cellcolor{yellow}0.968 & 30.41 & 50\\
 w/o contrastive & 30.76 & \cellcolor{pink}0.970 & \cellcolor{yellow}27.58 & 50\\
 pointnet & 30.75 & \cellcolor{pink}0.970 & 27.74 & 50\\
 w/o DAA & 30.78& \cellcolor{pink}0.970 & 27.83 & 50\\
 Top-3 & \cellcolor{yellow}30.80& \cellcolor{pink}0.970& 27.73& 50\\
 Top-5 & 30.79& \cellcolor{pink}0.970& 27.74& 50\\
 Top-10 &30.76 & \cellcolor{pink}0.970& 27.82& 50\\
 Full model & \cellcolor{pink}30.81 & \cellcolor{pink}0.970 & \cellcolor{pink}27.48 & 50\\
 \hline
 w/o con & 30.33 & \cellcolor{yellow}0.968 & \cellcolor{yellow}29.96 & 50\\
 w/o DAA & 30.42& \cellcolor{yellow}0.968 & 30.17 & 50\\
 Top-3 & \cellcolor{yellow}30.44& \cellcolor{pink}0.969& 29.98& 50\\
 Top-5 & 30.43& \cellcolor{yellow}0.968& 30.19& 50\\
 Full model & \cellcolor{pink}30.48& \cellcolor{pink}0.969 & \cellcolor{pink}29.94 & 50\\
 \bottomrule
 \end{tabular}
\end{table}

\begin{figure}[t]
  \centering
    \includegraphics[width=\linewidth]{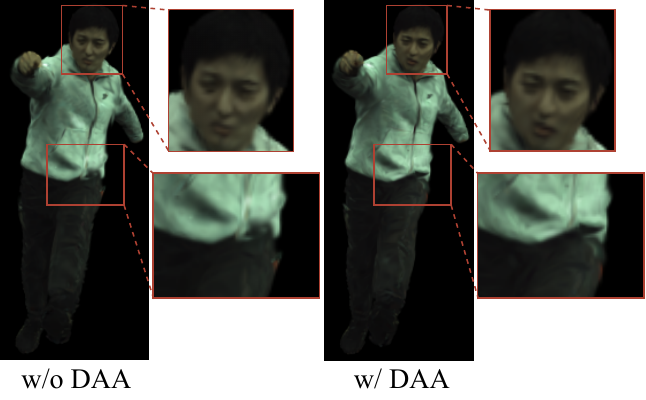}\\
  \caption{\textbf{Ablation Study} on DAA, which enhances multi-view 3D consistency, hence improving the overall rendering quality.}
  \label{fig:ablation_adj}
\end{figure}

\begin{figure}[t]
  \centering
    \includegraphics[width=\linewidth]{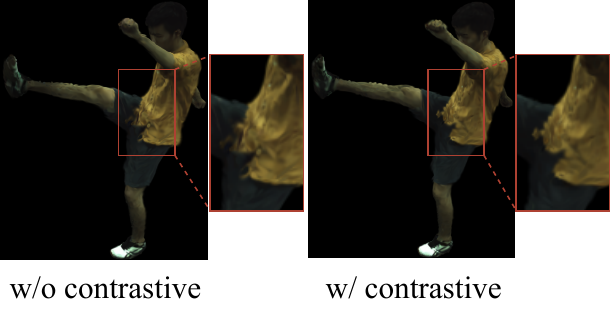}\\
  \caption{\textbf{Ablation Study} on 3D geometry contrastive learning, which removes the artifacts on highly articulated poses..}
  \label{fig:ablation_contr}
\end{figure}


\subsection{Ablation Study}
\label{sec:ablation}

In this section, we conduct ablation experiments using the ZJU-MoCap~\cite{peng2020neural} dataset with both full inputs and sparse inputs to evaluate the effectiveness of our proposed modules. We also conduct experiments to evaluate different backbones in 3D geometry contrastive learning. \textit{Notably, our \our maintains 3D consistency effectively without increasing any extra inference time.}



\medskip
\noindent
\textbf{Non-rigid deformation.} Non-rigid Deformation is designed for complex deformations, such as cloth bending and stretching. As shown in Tab.~\ref{tab:ablation}, non-rigid deformation is required to achieve optimal performance, demonstrating non-rigid regions are well rendered.

\medskip
\noindent
\textbf{Dynamic Avatar Adjustment.}
As shown in Tab.\ref{tab:ablation}, incorporating DAA results in our full model outperforming the baseline in terms of LPIPS which is particularly informative compared to other metrics in our setting~\cite{qian20243dgs,yang2024gaussianobject}. Fig.~\ref{fig:ablation_adj} illustrates that DAA serves as an effective 2D image supervision for 3D human body modeling, enhancing 3D consistency and reducing artifacts while improving multi-view consistency. Additionally, DAA reduces artifacts and inaccuracies in the geometry caused by sparse inputs, further enhancing the robustness and visual realism of the model in practical applications.

\medskip
\noindent
\textbf{Similar (pose/image) selection.} Our method selects a similar (pose/image) to supervise the generated avatar as described in Section~\ref{sec:daa}. Ablation studies in Tab.~\ref{tab:ablation} show that even when selecting less similar (poses/images), our approach significantly improves performance. Top-n refers to choosing the n-th most similar pose/image. These results highlight \our’s effectiveness in handling limited data or irregular large-scale human movements.

\medskip
\noindent
\textbf{3D geometry contrastive learning.} Our core idea is that cross-3D-modal contrastive learning can facilitate communication between 3D models for obtaining powerful representations. To verify this, we further do ablation studies to show qualitative comparisons in Tab.~\ref{tab:ablation} and Fig.~\ref{fig:ablation_contr}. We can find the full model (w/ contrastive) preserves finer details and provides
a more realistic and detailed reconstruction of clothing, demonstrating that 3D geometry contrastive learning enhances 3D consistency.

\medskip
\noindent
\textbf{Backbone for 3D contrastive learning.} In Tab.~\ref{tab:ablation}, we show the ablation study on different backbones, including PointNet~\cite{qi2017pointnet} (pointnet) and DGCNN~\cite{wang2019dynamic} (full model). This indicates that DGCNN captures local geometric details better by building a dynamic graph structure, while PointNet relies on global feature learning, which may cause some local information to be missing.


\section{Conclusion}

In this paper, we present \our, a 3D-consistent human modeling framework utilizing Gaussian Splatting with both full and sparse inputs. We first integrates a skeleton-driven rigid deformation and a non-rigid cloth dynamics deformation to create human avatar. To improve 3D consistency under sparse inputs, we use the intrinsic 3D consistency of images across poses. For each training image, we select similar pose/image from the dataset and adjust the deformed Gaussians to selected pose by Dynamic Avatar Adjustment (DAA). Minimizing the difference between the image rendered by adjusted Gaussians and image paired with selected similar pose serves as an additional supervision, hence enhancing the 3D consistency of human avatars. Furthermore, to enforce global 3D consistency across different representations of the same pose, we propose a 3D geometry contrastive learning. Extensive experiments on two popular datasets demonstrate that \our not only achieves superior fidelity in generating human avatars compared to current SOTA methods but also excels in handling both monocular and sparse input scenarios.  We hope that our method could foster further research in high-quality clothed human avatar synthesis from monocular views.

\medskip
\noindent
\textbf{Limitations.} 
1). \our lacks the capability to extract 3D meshes. Developing a method to extract meshes from 3D Gaussians is an important direction for future research. 
2). \our performs less effectively in reconstructing humans with complex clothing, as similar poses may still exhibit substantial differences in non-rigid deformations. This remains a challenge we plan to address in future work.

{\small
\bibliographystyle{ieee_fullname}
\bibliography{egbib}
}

\end{document}